\documentclass[lettersize,journal]{IEEEtran}
\usepackage{color}    
\usepackage{multirow}
\usepackage{amsmath}
\usepackage{amsfonts}
\usepackage{amssymb}
\usepackage{array}
\usepackage[caption=false,font=normalsize,labelfont=sf,textfont=sf]{subfig}
\usepackage{textcomp}
\usepackage{stfloats}
\usepackage{url}
\usepackage{verbatim}
\usepackage{graphicx}
\usepackage{color}
\usepackage{cite}
\usepackage{caption}
\usepackage{bbding}
\usepackage{amssymb}
\usepackage{algpseudocode}
\usepackage{algorithmicx}
\usepackage{algpseudocode}
\usepackage{algorithm}
\usepackage{booktabs}
\DeclareUnicodeCharacter{FF1A}{\textasciitilde}
\usepackage{makecell}
\usepackage[colorlinks,
            linkcolor=red,
            anchorcolor=blue,
            citecolor=green 
            ]{hyperref}

\usepackage{adjustbox}
\begin{document}

\title{Scaling Up Single Image Dehazing Algorithm by Cross-Data Vision Alignment for Richer Representation Learning and Beyond}

\author{Yukai Shi\dag, Zhipeng Weng\dag, Yupei Lin, Cidan Shi, Xingyuan Guo, Xiaojun Yang and Liang Lin, \IEEEmembership{Fellow, IEEE}

\thanks{
 {\dag} The first two authors share equal contribution.
 }

}



\maketitle

\begin{abstract}
In recent years, deep neural networks tasks have increasingly relied on high-quality image inputs. With the development of high-resolution representation learning, the task of image dehazing has received significant attention. Previously, many methods collect diverse image data for large-scale training to boost the performance on a target scene. Ignoring the domain gap between
different data, former de-hazing methods simply adopt multiple datasets for explicit large-scale training, which often makes the methods themselves be violated. To address this problem, we propose a novel method of cross-data vision alignment for richer representation learning to improve the existing dehazing methodology. Specifically, we call for the internal- and external knowledge should be further adapted with a self-supervised manner to fill up the domain gap. By using cross-data external alignment, the datasets inherit samples from different domains that are firmly aligned, making the model learn more robust and generalizable features. By using the internal augmentation method, the model can fully exploit local information within the images, and then obtaining more image details. To demonstrate the effectiveness of our proposed method, we conduct training on the Natural Image Dataset (NID). Experimental results show that our method clearly resolves the domain gap in different dehazing datasets and presents a new pipeline for large-scale training in the dehazing task. Our approach significantly outperforms other advanced methods in dehazing and produces dehazed images that are closest to real haze-free images. The code will be available at: \href{https://github.com/wengzp1/ScaleUpDehazing}{https://github.com/wengzp1/ScaleUpDehazing}
\end{abstract}


\section{Introduction}
\IEEEPARstart{I}{n} the field of natural images, haze usually leads to a decrease in image contrast and clarity. It makes the originally distinct boundaries of objects blurry, greatly affecting the visual effect of the image. Therefore, developing effective dehazing methods is of great significance for enhancing the application value of remote-sensing images. Previous dehazing methods mainly address hazy images by using atmospheric scattering models~\cite{dehazeformer1},~\cite{dehazeformer2},~\cite{dehazeformer3}. The model explains how light interacts with particulate matter in the atmosphere under hazy conditions, thereby affecting the imaging process of the images. The specific representation is as follows:
\begin{equation}
\label{eq1}
I(x)=J(x)t(x)\oplus A(1 - t(x))
\end{equation}
where $x$ is a pixel position in the image, $J(x)$ represents the true haze-free image, and $A$ is the global atmospheric light. $t(x)$ represents the medium transmission rate, which ranges from 0 to 1. Here, 0 indicates no transmission, meaning complete opacity. $I(x)$ represents the hazy image obtained after the haze-free image undergoes atmospheric scattering. 

Early dehazing methods utilize prior knowledge of the image to separately estimate the haze parameters $t(x)$ and $A$, and then further use the atmospheric scattering model to obtain the haze-free image $J(x)$. ALC~\cite{trinity9} proposes an image dehazing method that combines atmospheric light white balance correction, local light filtering, and high-altitude perspective priors to restore clarity to hazy images. While these methods are intuitive, they struggle to effectively remove haze in non-uniformly hazy images. 

Recently, many methods have emerged for use in the field of image dehazing. MSPD-Net~\cite{tim1} proposes a progressively optimized multi-stage dehazing network. It uses a physics-based feature enhancement block and selective kernel feature fusion to achieve image dehazing. MDFEN~\cite{tim2} proposes an enhanced network that fuses multiscale depth information. This network integrates multiscale features from depth maps and hazy images. It also incorporates a context attention mechanism to improve dehazing performance in scenes with deep depth of field and heavy fog. Oval-Net~\cite{tim3} proposes an end-to-end dehazing network that incorporates spatial and channel attention mechanisms. The network efficiently processes image data within an encoder-decoder structure, improving dehazing performance without relying on the atmospheric scattering model. TOENet~\cite{tim4} proposes an enhancement network designed to improve image quality in low-visibility environments. It uses a multilayer perceptron-based channel correlation extraction module and an encoder-decoder architecture to achieve image dehazing and enhancement in hazy and sandstorm conditions. Cycle-SNSPGAN~\cite{trinity56} improves the quality of dehazed images by combining spectral normalization and patch discriminator techniques.  
In addition, deep learning methods train on a large number of hazy and haze-free image pairs, and use models such as Transformers to learn the mapping from hazy images to clear ones. Dehamer~\cite{trinity21} introduces a transmission-aware 3D positional embedding module into the Transformer, providing relative positional information while incorporating priors related to haze density.

\begin{figure*}[!t]
\centering
\includegraphics[width=0.85\linewidth]{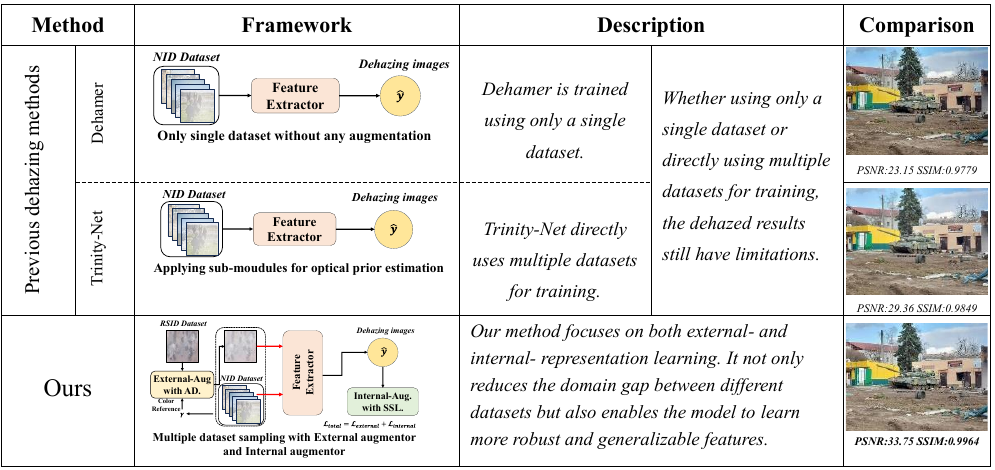}
\caption{Comparison with state-of-the-arts in terms of flowchart and characteristics. With the rapid development of Transformer~\cite{dehazeformerself,swintransformer30}, the demand for data has grown at the same time. However, the simple combination of multiple datasets achieves unsatisfactory results. \emph{To address this issue, we call for the internal- and external knowledge should be further augmented by vision alignment and self-supervised learning to perform an effective large-scale training.} }
\label{fig0}
\end{figure*}

However, the aforementioned methods mostly consider the model perspective and rarely take the data itself aspect into account. 
With the rapid development of Transformer~\cite{dehazeformerself,swintransformer30}, the demand for data has grown at the same time. Ignoring the domain gap between different data, former de-hazing methods usually adopt multiple datasets to a large model (e.g., Transformers), which often achieves unsatisfactory results. As shown in Fig.~\ref{fig0}, the Dehamer~\cite{dehazeformerself} applies the narrow training data to train a Transformer. Specifically, the training dataset lacks diversity in terms of scenes and types of haze, resulting in the model's inability to achieve good robustness. Moreover, Trinity-Net~\cite{trinityself} ignores the domain gap between existing datasets, and directly using multiple datasets for training, which often results in limited performance improvement. 

Then, we compare existing self-supervised dehazing methods with our method. SLAdehazing~\cite{selfdehaze1} employs a self-supervised adaptation strategy, leveraging contrastive learning directly on a large-scale synthetic dataset of hazy-clean image pairs. In contrast, our method enhances external data through cross-dataset alignment, significantly improving the model's robustness. Instead of guiding dehazing with depth information, as in DP-dehazing~\cite{selfdehaze2}, our method utilizes gamma correction and Weak-to-Strong Augmentation method to align the distributions of the training data. In comparison to SGDRL~\cite{selfdehaze3}, which focuses on disentangling features for dehazing, our method integrates cross-data alignment with self-supervision, emphasizing the robustness gained from training on diverse augmented data.

To address the data scarcity and uniqueness in large-scale training, we propose a novel method of external- and internal- knowledge alignment and augmentation, to improve the dehazing model. Specifically, we first expand the dataset externally using a knowledge alignment method. The datasets inherit samples from different domains that are firmly aligned, making the model learn more robust and generalizable features. Next, we apply weak-to-strong self-supervised learning methods to learn richer representations. Additionally, the dehazing results produced by our method visually appear closer to real haze-free images. To this end, the contributions of this paper can be summarized as follows:
\begin{itemize}

\item[$\bullet$]We expand the dataset externally using a knowledge alignment method. The dataset inherits samples from different domains that are aligned by the external data augmenter, making the model learn more robust and generalizable features.

\item[$\bullet$]We propose a Weak-to-Strong Augmentation method with self-supervised manner to learn richer representations. Compared with previous approaches, the proposed method learns more feature representations with spatial-invariant strategies and self-supervised learning.  

\end{itemize}

\begin{figure*}[!t]
\centering
\includegraphics[width=0.85\linewidth]{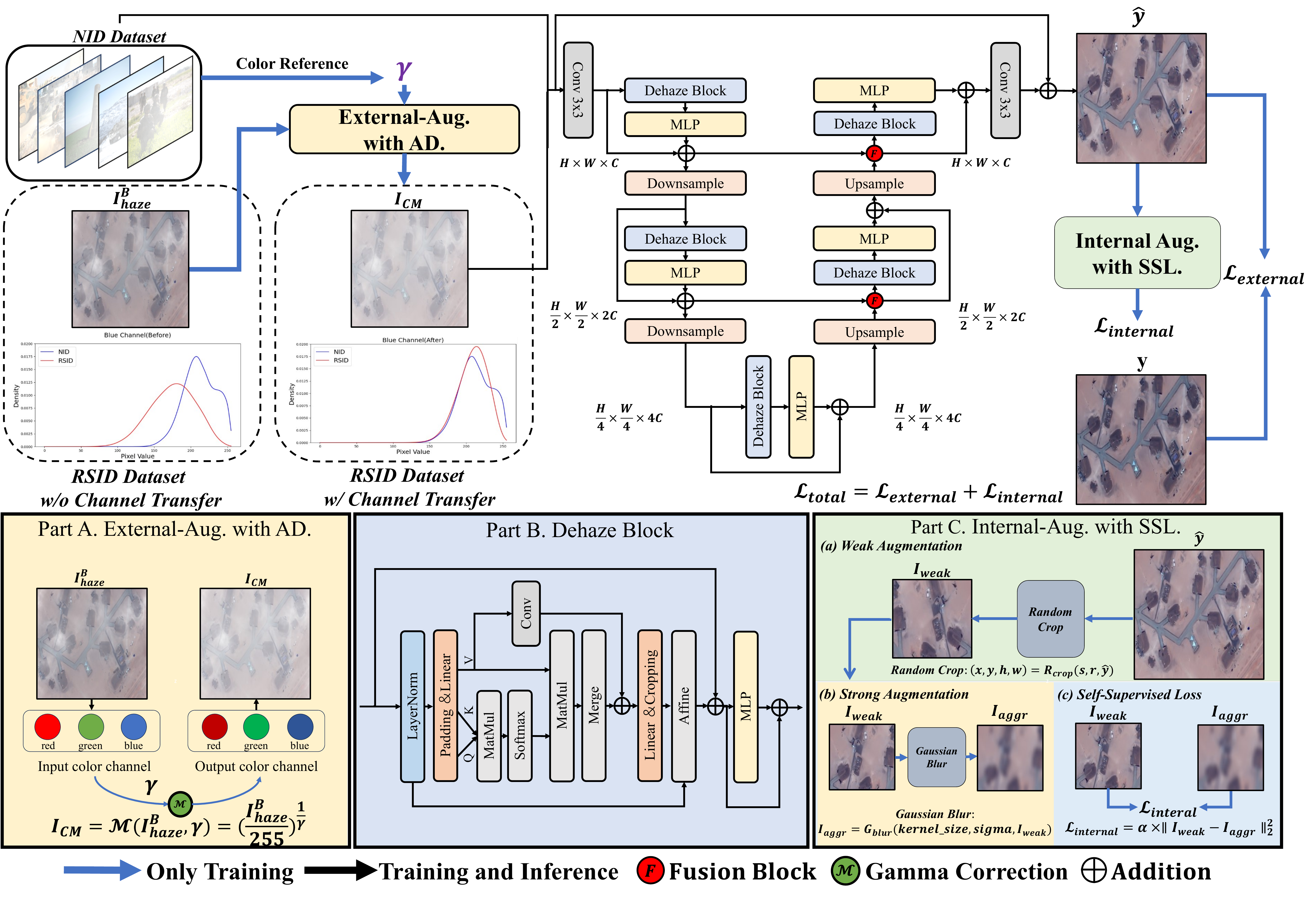}
\caption{Illustration of the proposed integrated framework for External-Augmentor with Auxiliary Datasets (External-Aug. with AD.) and Internal-Augmentor with Self-supervised Learning (Internal-Aug. with SSL.). In our method, we perform external augmentation on the target dataset to increase its diversity. The external augmentor makes the auxiliary dataset be aligned with the target dataset in the color channels. And gamma correction mainly affects the brightness and contrast of the images. Then, we apply the self-supervised learning method to augment the data internally. Specifically, the internal augmentor emphasizes attention to fine-grained details in images. By simultaneously augmenting the data externally and internally, our model achieves higher performance and robustness.}
\label{fig1}
\end{figure*}

\section{RELATED WORK}
\label{related work}
\subsection{Image Dehazing}
In the fields of natural image processing, image dehazing techniques play an important role. Hazy images are not conducive to the related tasks of vision, thus causing a lot of challenges to a certain extent. Recently, with the development of deep learning techniques, deep learning-based dehazing methods have gradually become a research hotspot. MSPD-Net~\cite{tim1} introduces a multistage progressive dehazing network. It achieves effective image dehazing using a physical model-based feature enhancement block and selective kernel feature fusion. MDFEN~\cite{tim2} proposes a multiscale depth fusion enhancement network. Oval-net~\cite{tim3} introduces an end-to-end dehazing network based on spatial and channel attention mechanisms. TOENet~\cite{tim4} introduces a two-in-one low-visibility enhancement network. FSE-DIF~\cite{tim5} introduces a fast single-image dehazing algorithm based on artificial multi-exposure image fusion. HECB~\cite{tim6} introduces a contrast enhancement algorithm based on the atmospheric scattering model. DehazeNet~\cite{dehazeformer8} utilizes the deep architecture of CNN to output the medium transmission map. Then, it uses the atmospheric scattering model to recover the haze-free image. Liu \emph{et al}~\cite{liu2023data} first employ a gamma correction on non-homogeneous dehazing task, which achieves a impressive results and gradually become a basic operation in dehazing task. Dehamer~\cite{trinity21} introduces an three-dimensional positional embedding module into the Transformer, incorporating priors related to haze density to estimate the haze density in different spatial regions. Previous methods ignore the data scarcity and uniqueness in large-scale training, which often results in limited performance improvement.  

\subsection{Self-supervised Learning}
Self-supervised learning(SSL) has achieved significant success in the field of computer vision~\cite{MSR19,MSR56,MSR37,MSR47,MSR5,MSR11,MSR57,shi2024diff,lu2024sirst}. It trains models by generating labels automatically, without the need for manually annotated datasets. The key idea of self-supervised learning is to utilize the inherent structure and relationships within the data itself to learn effective feature representations. Self-supervised learning can be trained on large amounts of cheap data, which is highly valuable in fields like image dehazing and natural image dehazing. Recently, many self-supervised learning-based methods have been applied to dehazing tasks. SLAdehazing~\cite{selfdehaze1} removes haze by pre-training on synthetic hazy-clean image pairs and adapting to real hazy images through contrastive learning. DP-dehazing~\cite{selfdehaze2} designs a depth-guided hybrid attention Transformer module, leveraging depth information to enhance dehazing performance. SGDRL~\cite{selfdehaze3} achieves unsupervised and semi-supervised single image dehazing through multi-level progressive feature disentanglement, a self-guided attention mechanism, and a disentanglement-guided module. In image dehazing tasks, self-supervised learning often need to be designed specifically for the characteristics of images under hazy conditions. To address this issue, our method leverages a generalized and effective self-supervised learning approach to enhance the model's capability in processing hazy images.

\section{Methodology}
As shown in Fig.~\ref{fig1}, our proposed method consists of two parts: external data augmentor and internal data augmentor. In Fig.~\ref{fig1} , N, C, H, and W represent the batch size, the number of channels, the height, and the width of the images, respectively. Our method utilizes the NID~\cite{trinityself} and RSID~\cite{trinityself}, which consist of hazy images and their corresponding clear images. These datasets are used for both training and evaluating the effectiveness of the proposed method. Firstly, we propose an external- augmentation method to reduce the contrast degradation between different hazy datasets. External data augmentor makes the augmented dataset approximate the target dataset in the color space. The augmented samples obtained by our method can better match the distribution of the target domain. We believe that dehazing networks trained on a larger and more consistent set of samples are more likely to achieve superior performance.

Building upon the augmentation of different domain datasets externally, we further explore additional locally effective information by augmenting the dataset internally. The model becomes more robust by coordinating external and internal augmentation. For internal- augmentation, we introduce a weak-to-strong coordinated data augmentation self-supervised learning method to improve the local details of images. The model can focus more on the details of the images by calculating the difference between weakly and strongly augmented images.

\begin{figure}[!t]
\centering
\includegraphics[width=0.99\linewidth]{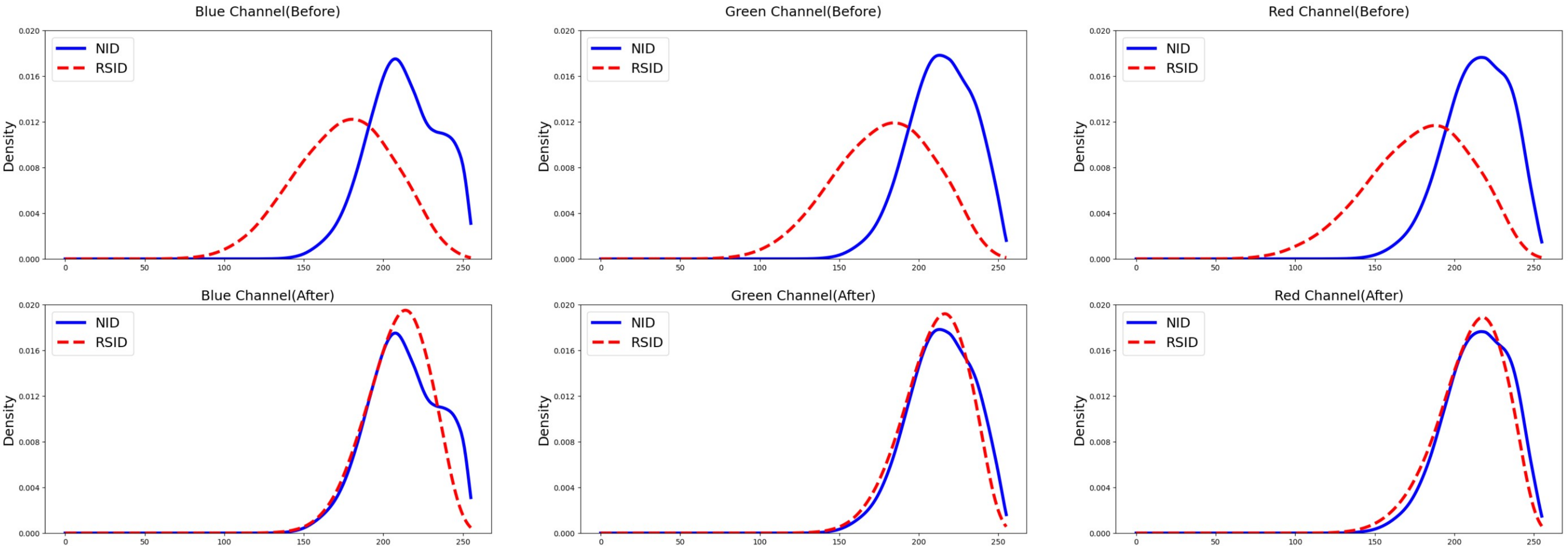}
\caption{A demonstration of domain gap between different dahazing datasets. We perform gamma correction on the RSID based on the RGB values of the NID. It can be observed that before correction, there is a significant difference between the distributions of the datasets in the RGB channels. After external- augmentation, the distributions of RSID images are closely aligned with those of NID images in the RGB channels.}
\label{fig2}
\end{figure}
\subsection{Cross-data External Alignment}

\begin{figure}[t]
\centering
\includegraphics[width=0.90\columnwidth]{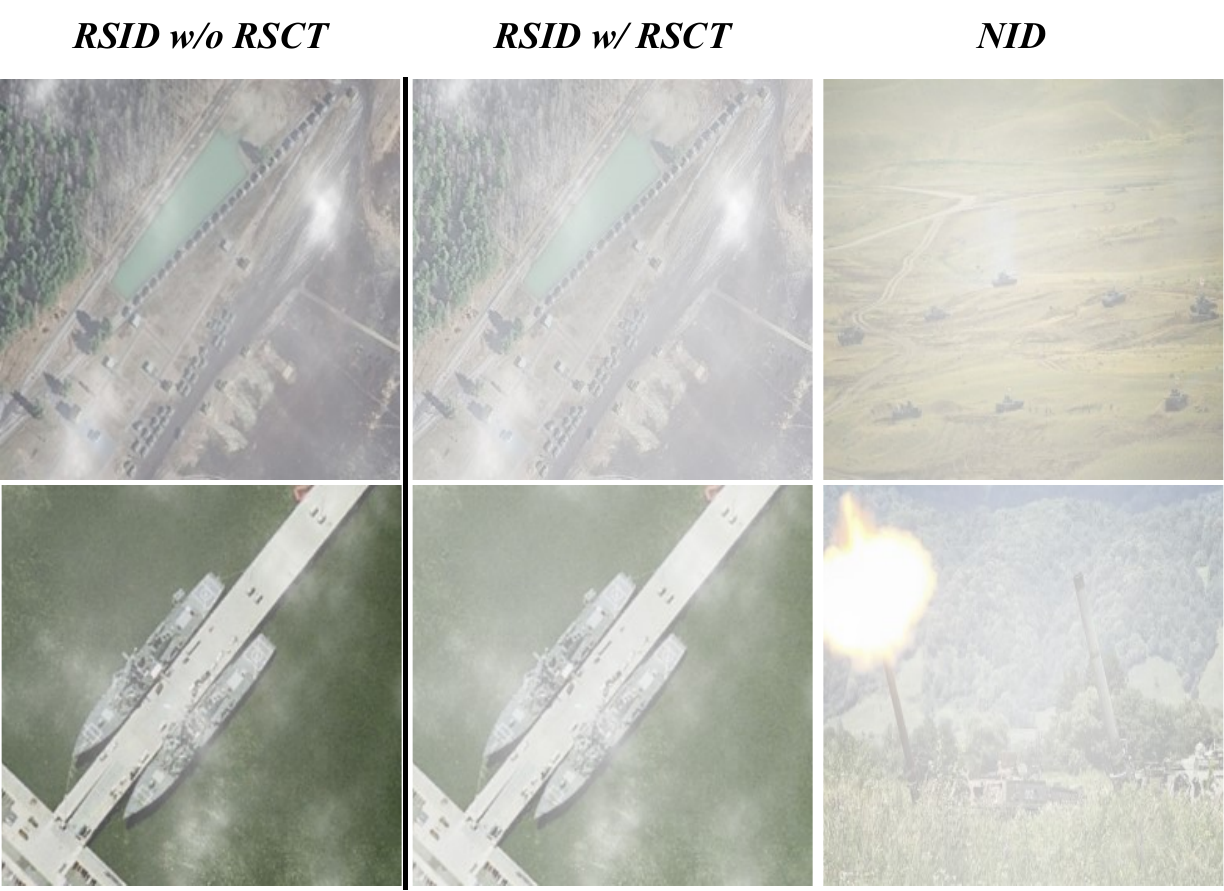}
\caption{Illustration of external augmentation. The RSID images become closer to the NID dataset by external augmentor.}
\label{fig3}
\end{figure}

We present an external-augmentor to address the domain gap between different datasets. The proposed external-augmentor achieves efficient domain adaptation without an additional training phase. Specifically, we first calculate the gamma values $\gamma _{R,G,B}$ for the three channels to be transferred to the target domain. Then, we perform gamma calibration on the images to be transferred, as follows:
\begin{equation}
\label{eq2}
{O_{R,G,B} =\mathcal{M}(I_{R,G,B})=\left (  \frac{I_{R,G,B} }{255} \right ) ^{\frac{1}{ \gamma _{R,G,B} } }}   
\end{equation}
where $I_{R,G,B}$ and $O_{R,G,B}$ represent the input and output pixel intensities ([0,255]), respectively, and $\gamma$ is the gamma factor. The subscripts R, G, B represent the color channels, and the values of different channels are unique. Our method adjusts the RSID to make it more similar to the NID from color space. As shown in Fig.~\ref{fig2}, the RSID exhibit a closer alignment with the RGB distribution of the NID after external augmentation. By performing augmentation on the target dataset, our method effectively increases the diversity of dataset and enriches it with more varied examples, which helps improve the robustness. Some of the transformed data is shown in Fig.~\ref{fig3}. The first column shows the original RSID images, the second column shows the new RSID images after data preprocessing, and the third column shows the target dataset NID. As shown in Fig.~\ref{fig3}, the RSID images become closer to the NID after the external augmentor. To illustrate our method intuitively, we demonstrate the algorithmic flow of external augmentation. As shown in Algorithm~\ref{alg:external-augmentor}, we apply the RSID as an example to present the implementation of the channel correction. First, we calculate the average values of the three color channels (R, G, B) for both the source domain and the target domain, respectively. Then, we use the gamma correction formula to derive the required gamma value. Next, we apply the calculated gamma value to the source domain to adjust its three color channel values, making them approximate those of the target domain.

\begin{algorithm}[t!]
\caption{External-Augmentor for Cross-data Alignment}
\label{alg:external-augmentor}
\begin{algorithmic}

\State \textbf{Input:} $I_{{haze}}\in R^{N\times C\times H\times W} $ :RSID origin hazy images
\State \textbf{Intermediate parameter:} 
\State $\gamma_{R},\gamma_{G},\gamma_{B}$: gamma transform values for channel migration
 \State {\bf Output:} $I_{{CM}}\in R^{N\times C\times H\times W} $ :RSID hazy images after channel migration transformation

\\ \\
 $\triangleright$ Three channel color averages for the RSID origin hazy images

\For{$n\gets1$ to N}
    \State $Mean(I_{RSID-R})$ = $Mean(I_{RSID-R}^{n} )$
    \State $Mean(I_{RSID-G})$ = $Mean(I_{RSID-G}^{n} )$
    \State $Mean(I_{RSID-B})$ = $Mean(I_{RSID-B}^{n} )$
    
\EndFor

\\
\State $\triangleright$ Three channel color averages for the NID hazy images
\For{$n\gets1$ to N}
    \State $Mean(I_{NID-R})$ = $Mean(I_{NID-R}^{n} )$
    \State $Mean(I_{NID-G})$ = $Mean(I_{NID-G}^{n} )$
    \State $Mean(I_{NID-B})$ = $Mean(I_{NID-B}^{n} )$
    
\EndFor

\\
\State $\triangleright$ compute $\gamma_{R},\gamma_{G},\gamma_{B}$
\State $Mean(I_{NID-R})$ =$\left (\frac{Mean(I_{RSID-R})}{255} \right )^{\frac{1}{\gamma _{R} } } $
\State $Mean(I_{NID-G})$ =$\left (\frac{Mean(I_{RSID-G})}{255} \right )^{\frac{1}{\gamma _{G} } } $   
\State $Mean(I_{NID-B})$ =$\left (\frac{Mean(I_{RSID-B})}{255} \right )^{\frac{1}{\gamma _{B} } } $    

\\
\State $\triangleright$ Get $I_{{CM}}$
\For{$n\gets1$ to N}
     \State $O_{CM-R}$=$\left (\frac{I_{R}}{255} \right )^{\frac{1}{\gamma _{R} } }, O_{CM-G}$=$\left (\frac{I_{G}}{255} \right )^{\frac{1}{\gamma _{G} } }, O_{CM-B}$=$\left (\frac{I_{B}}{255} \right )^{\frac{1}{\gamma _{B} } }$
\EndFor
\\
\end{algorithmic}
\end{algorithm}

\begin{algorithm}[t!]
\caption{Internal-Augmentation with SSL}
\label{alg:internal-augmentor}
\begin{algorithmic}

\State \textbf{Input:}  $\hat{y}$:  the results of the enhanced network, $\alpha$=0.1
\State {\bf Output:} $\mathcal{L}_{internal}$:  internal augmentation loss
\\ 
\For{$i\gets1$ to I}
    \State$I_{weak}$=$P_{weak}(\hat{y} )$
    \State$I_{aggr}$=$P_{aggr}(I_{weak} )$
    \State$\mathcal{L}_{internal}$ =$\alpha$ ×$\|I_{weak}-I_{aggr}  \|_2^2$
    \State$\alpha$ = $\alpha \times \frac{1}{2} \left( \cos\left(\pi \frac{i}{I}\right) + 1 \right)$

\EndFor
\end{algorithmic}
\end{algorithm}

\subsection{Feature Representation Extraction}

As shown in Fig.~\ref{fig1}, the dehazing network is composed of a 5-layer U-Net structure~\cite{dehazeformerself}. The network includes two downsampling and upsampling steps. In the network, the input hazy image passes through Dehaze Block modules and feature fusion modules to finally obtain the output dehazed image $\widehat{y}$. As shown in part B of Fig.~\ref{fig1}, in the Dehaze Block module, the input feature maps undergo normalization and padding layers. Then they are projected into Q , K and V (query, key, value) using linear layers, used to calculate attention scores~\cite{swintransformer30}. Additionally, the Dehaze Block module applies multi-head self-attention within a window to handle relationships between image patches. It allows the dehazing network to consider correlations between different positions while extracting features. Thus, the model can learn structural information in the image more effectively to improve the model performance.

Different from directly integrating with the traditional U-Net network architecture, the feature fusion module learns channel attention weights of different feature maps through operations like global average pooling~\cite{dehazeformer45}. Thus, the network can dynamically adjust its contributions based on the importance of each feature map. In the feature fusion module, assuming there are two feature maps $f_1$ and $f_2$, We project $f_1$ to $f_1'$ using a linear layer. We then use operations such as $GAP(.)$ sequentially to obtain fusion weights $w_1$ and $w_2$.

Next, we use the weights to fuse $f_1'$ and $f_2$ with the following additional short residual:
\begin{equation}
\label{eq3}
x=w_1f_1'+w2f_2+f_2
\end{equation}

In the final short residual fusion stage, the two feature maps are fused through linear combination while introducing an additional short residual term. This design helps preserve the information from the original input while introducing the fused features. It allows the network to better utilize multi-scale and multi-level feature information, thus improving the accuracy and robustness of the dehazing effect.

Then, we use the L1 loss function to train the feature extractor, obtaining the loss for external data:
\begin{equation}
\label{eq4}
\mathcal{L}_{external}=|y-\widehat{y}  |
\end{equation}

\begin{figure}[!t]
\centering
\includegraphics[width=\linewidth]{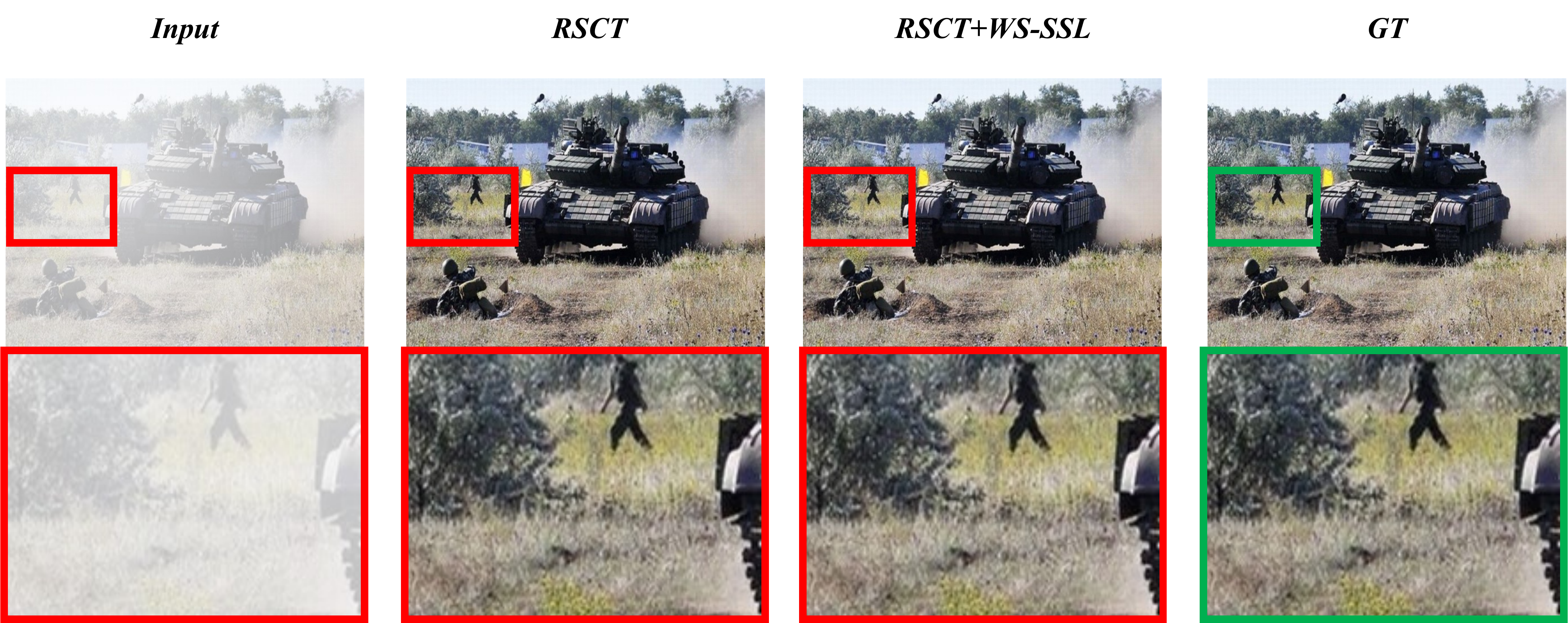}
\caption{Dehazed images obtained with and without internal data augmentor.}
\label{fig5}
\end{figure}

\subsection{Internal- Augmentation with Self-supervised Learning}

For the internal dataset, we propose an internal- augmentation strategy to improve the model's ability to preserve details and contrast in dehazed images. Building upon external dataset augmentation, it emphasizes attention to local details in images, making the model more robust. Specifically, we apply different degrees of data augmentation to the output results $\widehat{y}$ of the augmentation model. This is represented as follows:
\begin{equation}
\label{eq5}
I_{weak}=P_{weak}(\widehat{y} )
\end{equation}
where $P_{weak}()$ represents a weak data augmentation operation, and $I_{weak}$ denotes the image generated by performing weak augmentation on $\widehat{y}$. Subsequently, we apply strong augmentation to $I_{weak}$, which is specifically represented as follows:
\begin{equation}
\label{eq6}
I_{aggr}=P_{aggr}(I_{weak} )
\end{equation}
where $I_{aggr}$ is the image generated through strong augmentation based on $I_{weak}$, and $P_{aggr}$ represents the strong augmentation operation. Specifically, weak augmentation refers to random cropping, and strong augmentation $P_{aggr}$ is Gaussian blurring. In the Gaussian algorithm, our parameters are set as follows: the kernel size is set to 9, and the standard deviation is set to (0.1, 5). Finally, we compute the L2 norm between $I_{weak}$ and $I_{aggr}$, obtained through weak-strong augmentation to optimize the model's internal loss as:
\begin{equation}
\label{eq7}
\mathcal{L}_{internal} =\alpha  \times \|I_{weak}-I_{aggr}  \|_2^2
\end{equation}
\begin{equation}
\label{eq8}
\alpha = \alpha \times \frac{1}{2} \left( \cos\left(\pi \frac{s}{S}\right) + 1 \right) 
\end{equation}
where $\alpha$ is a hyperparameter that varies with cosine decay. In experiments, we set the initial value of $\alpha= 0.1$, $s$ is the current training iteration and $S$ is the total number of training iterations. As the number of training iterations increases, the cosine decay function ensures that the internal loss decreases along a smooth curve. It avoids sudden changes that could cause instability, ensuring the model's stability during the training process. By utilizing the self-supervised learning with $\mathcal{L}_{internal}$, we can obtain the internal similarity between data, which allows the model to learn richer features and be more robust.

\begin{figure*}[!t]
\centering
\includegraphics[width=\linewidth]{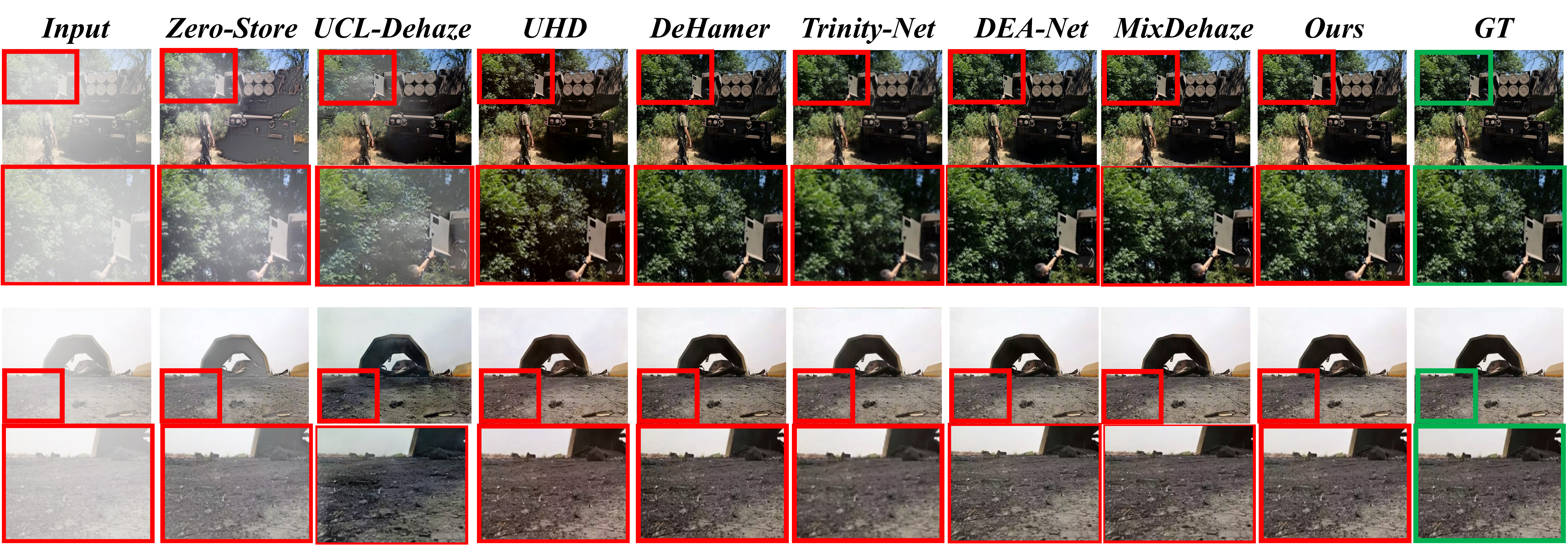}
\captionsetup{justification=centering}
\caption{Dehazed images obtained by state-of-the-art methods on the NID. Zooming up for better view. }
\label{fig6}
\end{figure*}

\begin{figure*}[t]
\centering
\includegraphics[width=0.99\linewidth]{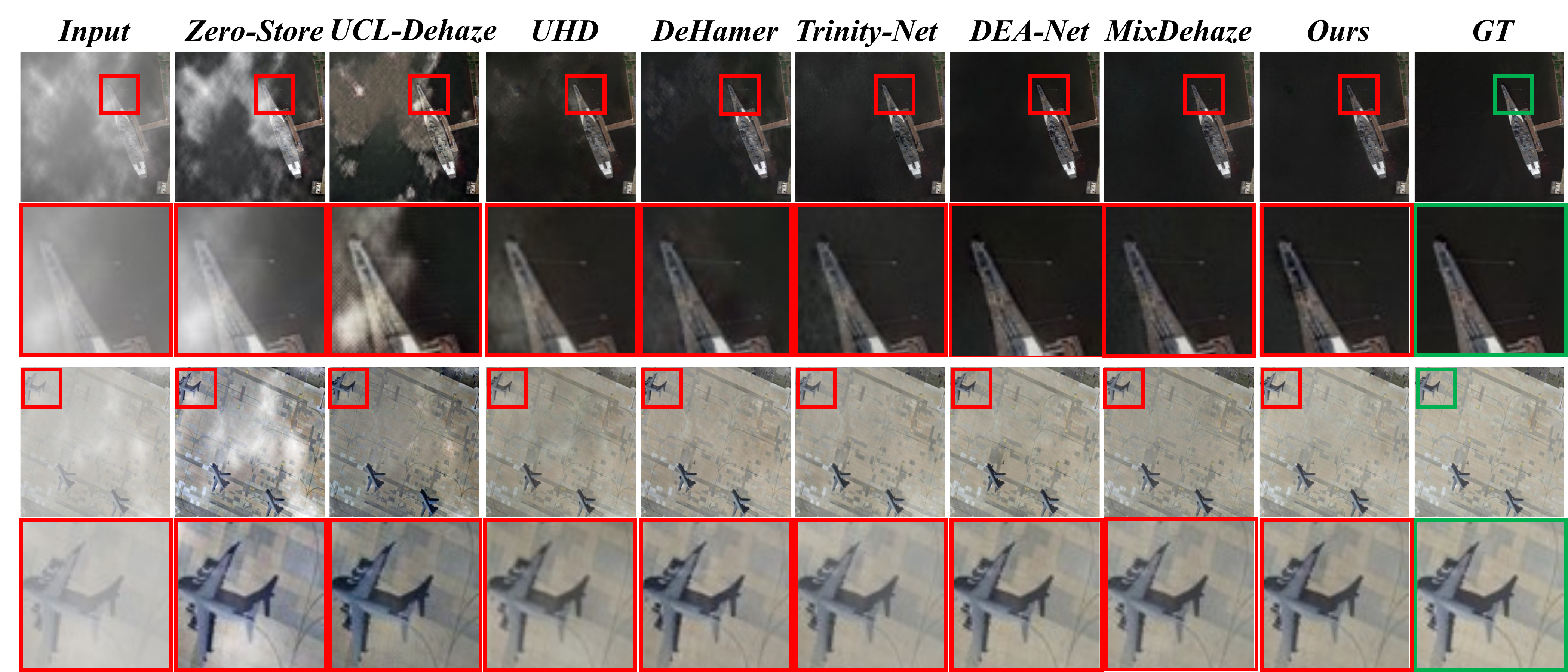}
\captionsetup{justification=centering}
\caption{Dehazed images obtained by state-of-the-art methods on the RSID. Zooming up for better view.}
\label{fig7}
\end{figure*}

\begin{figure*}[t]
\centering
\includegraphics[width=\linewidth]{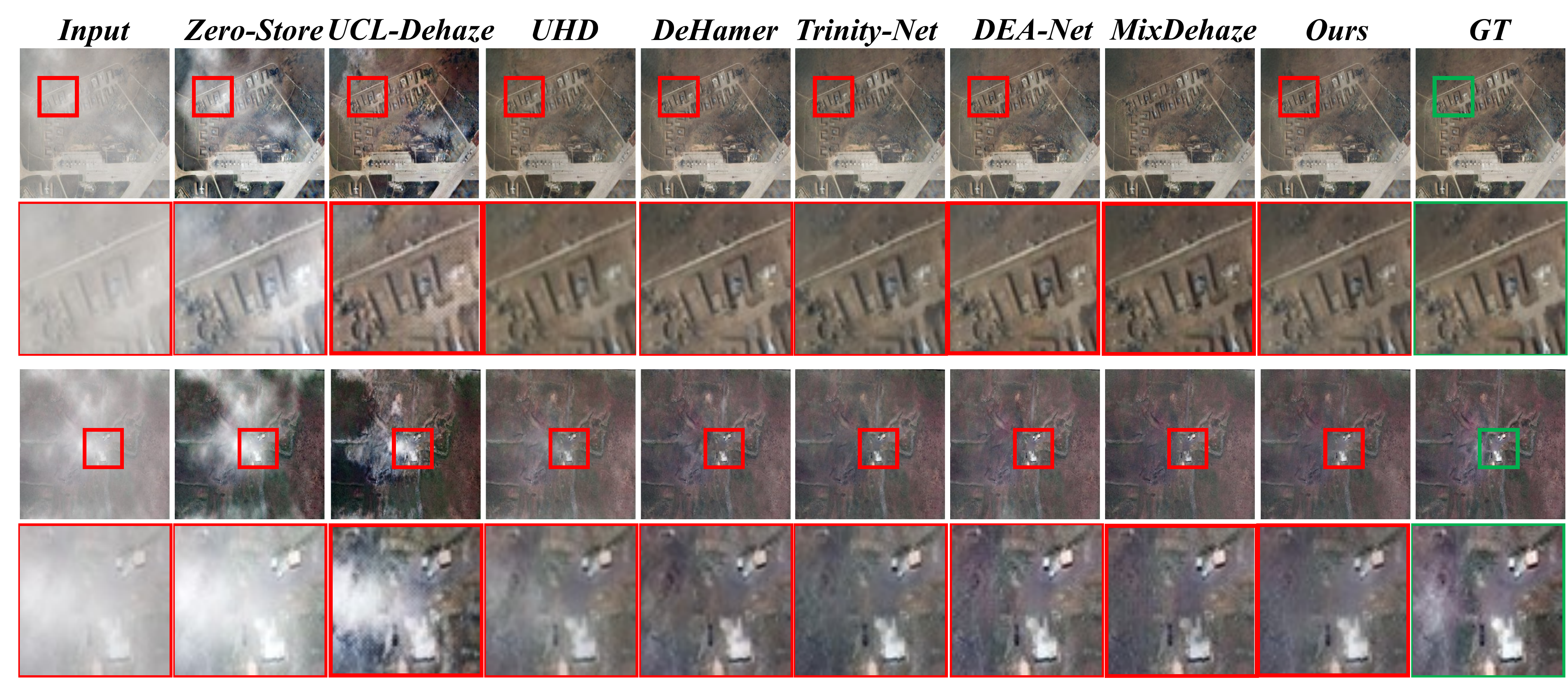}
\caption{Dehazed images obtained by state-of-the-art methods on the RSID. Zooming up for better view.}
\label{fig8}
\end{figure*}

\begin{figure}[!t]
\centering
\includegraphics[width=\linewidth]{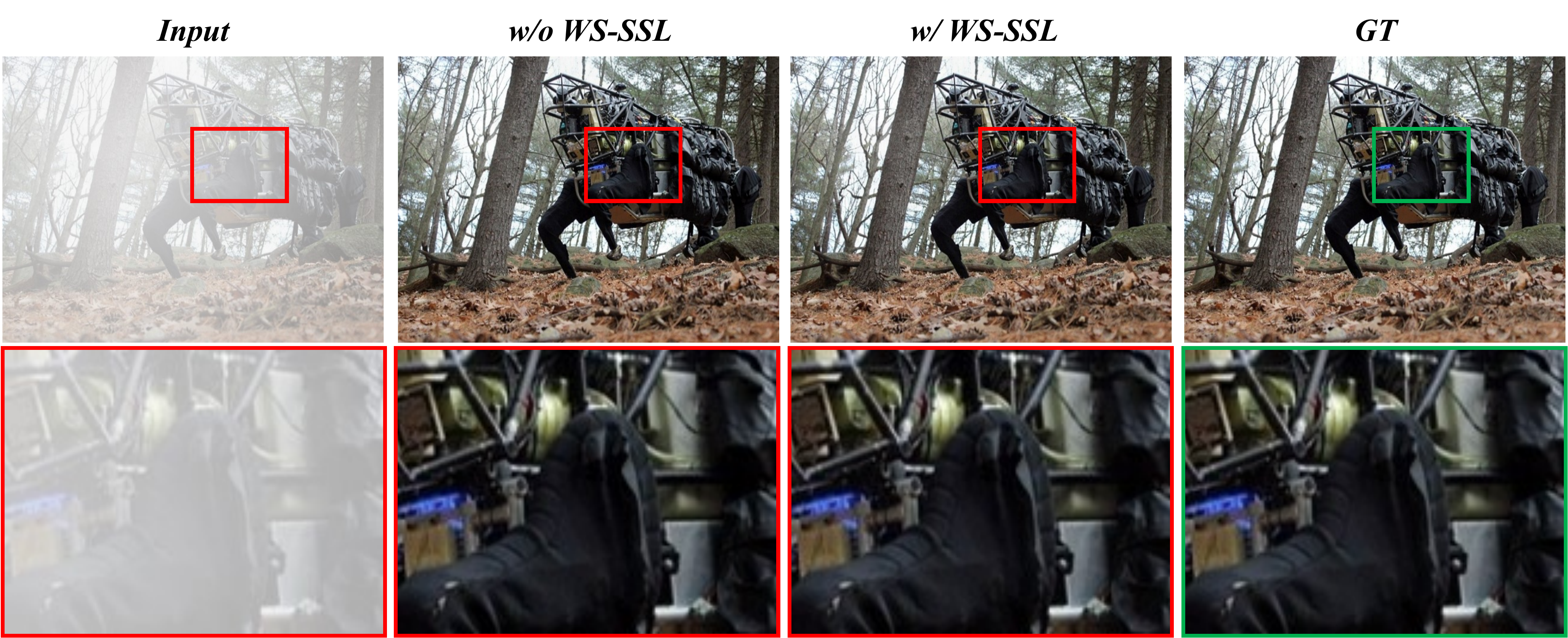}
\caption{Dehazed Images Obtained After Internal- Augmentation and Self-Supervised Module. }
\label{fig9}
\end{figure}

\begin{figure}[!t]
\centering
\includegraphics[width=\linewidth,height=0.4\textheight,keepaspectratio]{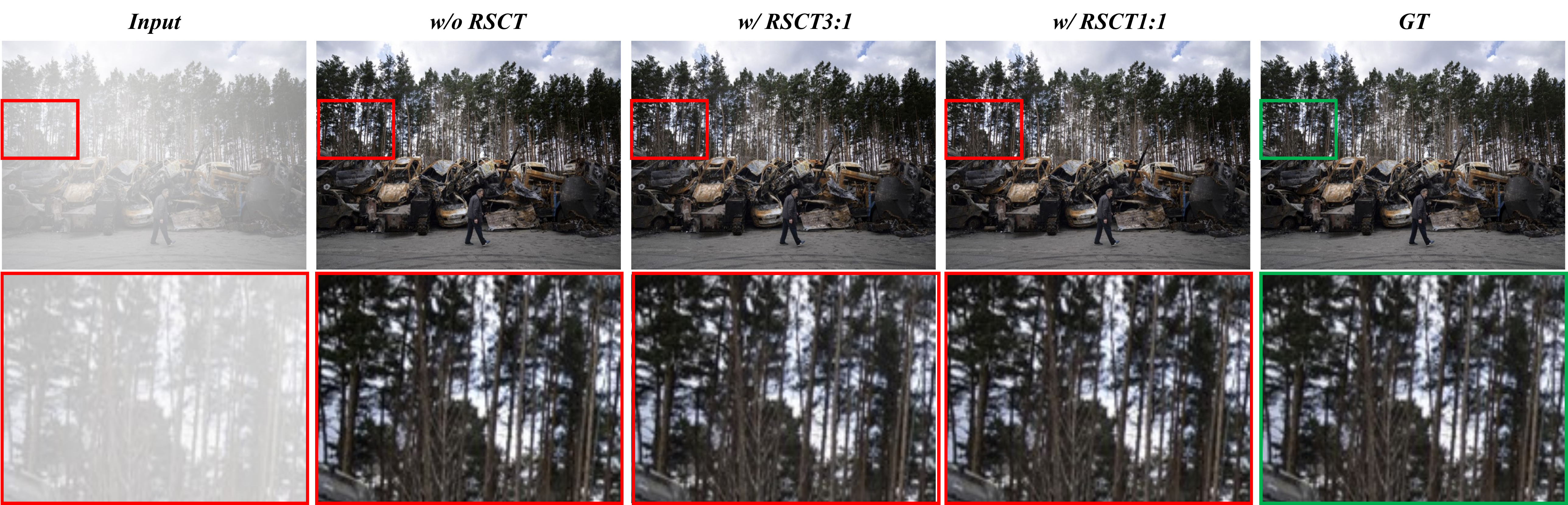}
\caption{Dehazed images obtained using external- augmentation method with ratios of 3:1 and 1:1 respectively.}
\label{fig10}
\end{figure}

\begin{figure}[!t]
\centering
\includegraphics[width=\linewidth,height=0.4\textheight,keepaspectratio]{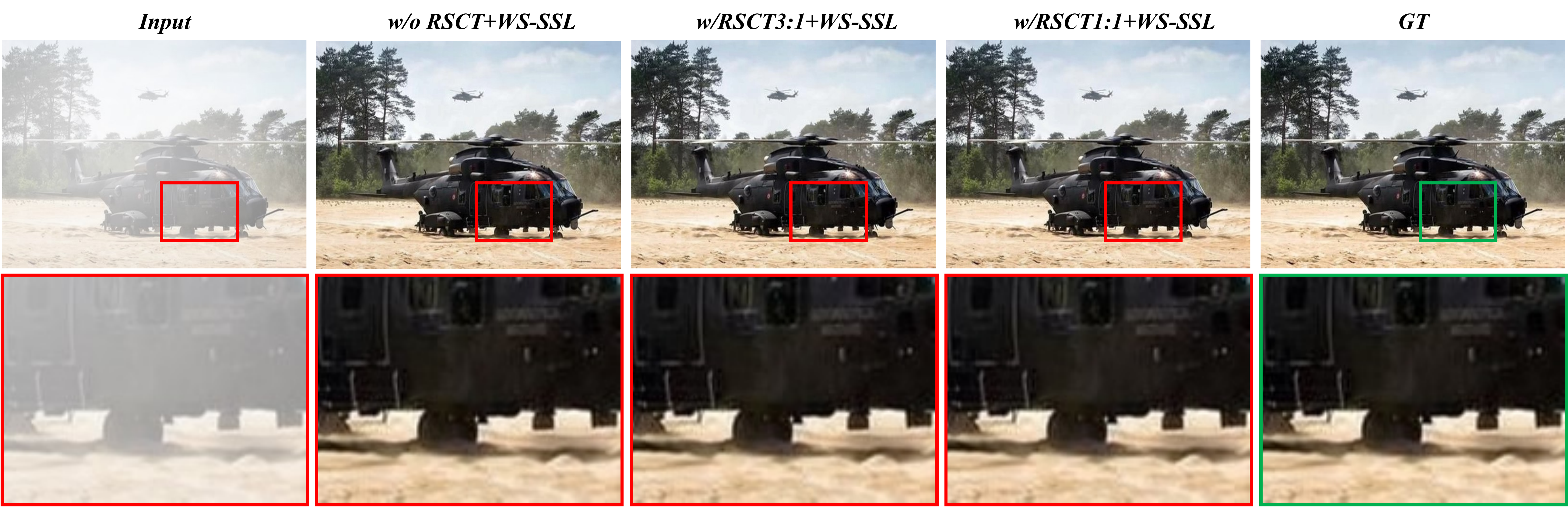}
\caption{Dehazed images obtained using both external and internal- augmentation methods. }
\label{fig11}
\end{figure}

\begin{figure}[ht]
\centering
\includegraphics[width=0.85\columnwidth]{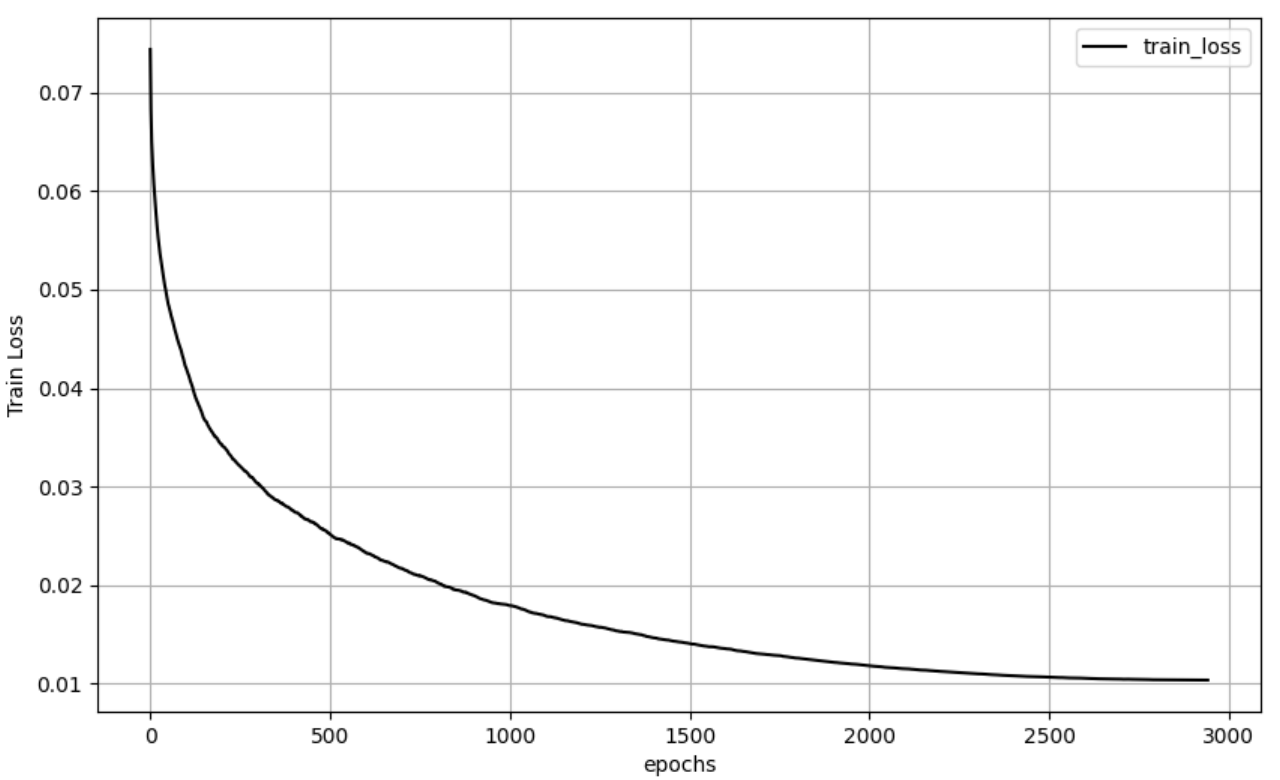}
\caption{The training loss curve of our method.}
\label{fig12}
\end{figure}

\begin{table}[ht]
    \centering
    \caption{Quantitative results on the NID.}
    \label{table1}
    \resizebox{\columnwidth}{!}{%
        \begin{tabular}{lcccc}
            \toprule
            Method & $PSNR\uparrow$ & $SSIM\uparrow$ & $MSE(\times 10^3)\downarrow$ & $FSIM\uparrow$\\
            \midrule
            Zero-Store~\cite{trinity18} & 16.09 & 0.7849 & 1.9159& 0.9550 \\
            UCL-Dehaze~\cite{ucldehaze} & 18.03 & 0.8830 & 2.3157 & 0.9349 \\
            DeHamer~\cite{trinity21} & 23.11 & 0.9369 & 0.3959 & 0.9738 \\
            UHD~\cite{trinity13} & 23.72 & 0.9182 & 0.2921 & 0.9700 \\
            Trinity-Net~\cite{trinityself} & 28.80 & 0.9743 & 0.2581 & 0.9906 \\
            DEA-Net~\cite{deanet} & 28.04 & 0.9814 & 0.3713 & 0.9944 \\
            MixDehazeNet~\cite{mixdehazenet} & 29.74 & 0.9880 & 0.2552 & 0.9957 \\
            Ours & \textcolor{red}{\textbf{31.14}} & \textcolor{red}{\textbf{0.9896}} & \textcolor{red}{\textbf{0.1973}}& \textcolor{red}{\textbf{0.9962}} \\
            \bottomrule
        \end{tabular}%
    }
    \label{table1}
\end{table}

\begin{table}[ht]
    \centering
    \caption{Quantitative results on the RSID.}
    \label{table2}
    \resizebox{\columnwidth}{!}{%
        \begin{tabular}{lcccc}
            \toprule
            Method & $PSNR\uparrow$ & $SSIM\uparrow$ & $MSE(\times 10^3)\downarrow$ & $FSIM\uparrow$\\
            \midrule
            Zero-Store~\cite{trinity18} & 16.65 & 0.7173&2.0185&0.8904 \\
            {UCL-Dehaze}~\cite{ucldehaze} & 17.59 & 0.8090&2.4816&0.8739 \\
            UHD~\cite{trinity13} & 22.06 & 0.8891&1.7286&0.9421 \\
            DeHamer~\cite{trinity21} & 23.61 & 0.8990&1.1223&0.9475 \\
            Trinity-Net~\cite{trinityself} & 23.78 & 0.9224&1.0805&0.9541 \\
            MixDehazeNet~\cite{mixdehazenet} & 24.20 & 0.9238&0.9422&0.9591 \\
            DEA-Net~\cite{deanet} & 24.79 & 0.9343&0.8787&0.9646 \\
            Ours & \textcolor{red}{\textbf{25.44}} & \textcolor{red}{\textbf{0.9399}}&\textcolor{red}{\textbf{0.7481}}&\textcolor{red}{\textbf{0.9665}} \\
            \bottomrule
        \end{tabular}%
    }
    \label{table2}
\end{table}

\begin{table}[ht]
    \caption{Efficiency analysis.}
    \label{table6}
    \centering
    \small
    \resizebox{\columnwidth}{!}
    {
    \begin{tabular}{|c|c|c|c|c|}
    \hline
   Model & UHD~\cite{trinity13}        & Trinity~\cite{trinityself} & DEA-Net~\cite{deanet} &  Ours                                  \\ \hline 
    Params (M)                  & 34.54    & 20.24 &  7.79  & \textbf{2.51}                                \\ \hline
    {GFLOPS}                  & 106    & 144 &  \textbf{92}  & 110                              \\ \hline
    {Runtime(hours)}                  & 11    & \textbf{12}  &  34 & 40                                \\ \hline
    $PSNR$                   & 23.72   & 28.80  & 28.04   & \textbf{31.14}  \\ \hline
    $SSIM$                    & 0.9182   & 0.9743                       & 0.9814  & \textbf{0.9896} \\ \hline

    \end{tabular}
    }
    
\end{table}

\begin{table}[ht]
    \centering
    \caption{Ablation study using internal-augmentation and self-supervised module.}
    \label{table3}
    \resizebox{0.75\columnwidth}{!}{%
        \begin{tabular}{|l|c|c|}
            \hline
            \multirow{2}{*}{\begin{tabular}[c]{@{}c@{}}Internal-augmentor\\ WS-SSL\end{tabular}} & \multicolumn{2}{c|}{N100} \\
            \cline{2-3}
            & $PSNR\uparrow$ & $SSIM\uparrow$ \\
            \hline
            \multicolumn{1}{|c|}{$\times$} & 30.06 & 0.9871 \\
            \hline
            \multicolumn{1}{|c|}{$\checkmark$} & 30.20 & 0.9875 \\
            \hline
        \end{tabular}%
    }
\end{table}

\begin{table}[ht]
    \centering
    \caption{Ablation study using only external-augmentation.}
    \label{table4}
    \resizebox{0.9\columnwidth}{!}{%
        \begin{tabular}{|c|c|c|c|}
            \hline
            \multicolumn{2}{|c|}{External-augmentor} & \multicolumn{2}{c|}{{N100}} \\
            \hline
            RSCT3:1 & RSCT1:1 & $PSNR\uparrow$ & $SSIM\uparrow$ \\
            \hline
            $\times$ & $\times$ & 30.06 & 0.9871 \\
            \hline
            $\checkmark$ & $\times$ & 30.47 & 0.9878 \\
            \hline
            $\times$ & $\checkmark$ & 30.71 & 0.9892 \\
            \hline
        \end{tabular}%
    }
\end{table}

\begin{table}[ht]
    \centering
    \caption{Ablation study using both external and internal-augmentation methods.}
    \label{table5}
    \renewcommand{\arraystretch}{2} 
    \large 
    \resizebox{0.85\columnwidth}{!}{%
        \begin{tabular}{|>{\centering\arraybackslash}p{2.2cm}|>{\centering\arraybackslash}p{2.2cm}|>{\centering\arraybackslash}p{2.2cm}|c|c|}
            \hline
            \multicolumn{3}{|c|}{External-augmentor and Internal-augmentor} & \multicolumn{2}{c|}{N100} \\
            \hline
            WS-SSL & RSCT3:1 & RSCT1:1 & $PSNR\uparrow$ & $SSIM\uparrow$ \\
            \hline
            $\times$ & $\times$ & $\times$ & 30.06 & 0.9871 \\
            \hline
            $\checkmark$ & $\checkmark$ & $\times$ & 30.70 & 0.9896 \\
            \hline
            $\checkmark$ & $\times$ & $\checkmark$ & \textbf{31.14} & \textbf{0.9896} \\
            \hline
        \end{tabular}%
    }
\end{table}

\begin{table}[ht]
    \centering
    \caption{Ablation study using our method toward Trinity-Net on the NID.}
    \label{table7}
    \resizebox{0.9\columnwidth}{!}{%
        \begin{tabular}{|c|c|c|c|}
            \hline
            \multicolumn{2}{|c|}{Method} & $PSNR\uparrow$ & $SSIM\uparrow$ \\
            \hline
            \multicolumn{2}{|c|}{Trinity-Net+origion DA} & 28.80 & 0.9743 \\
            \hline
            \multicolumn{2}{|c|}{Trinity-Net+our method} & 29.12 & 0.9755 \\
            \hline
        \end{tabular}%
    }
\end{table}

\begin{table}[ht]
    \centering
    \caption{Ablation study using our method in DEA-Net on the RSID.}
    \label{table8}
    \resizebox{0.9\columnwidth}{!}{%
        \begin{tabular}{|c|c|c|c|}
            \hline
            \multicolumn{2}{|c|}{Method} & $PSNR\uparrow$ & $SSIM\uparrow$ \\
            \hline
            \multicolumn{2}{|c|}{DEA-Net+origion DA} & 24.79 & 0.9343 \\
            \hline
            \multicolumn{2}{|c|}{DEA-Net+our method} & 25.28 & 0.9382 \\
            \hline
        \end{tabular}%
    }
\end{table}

\begin{table}[ht]
    \centering
    \caption{\textcolor{red}{Ablation study on our method and fixed label supervision.}}
    \label{table9}
    \resizebox{0.9\columnwidth}{!}{%
        \begin{tabular}{|c|c|c|c|}
            \hline
            \multicolumn{2}{|c|}{Method} & $PSNR\uparrow$ & $SSIM\uparrow$ \\
            \hline
            \multicolumn{2}{|c|}{Fixed Label Supervision} & 30.86 & 0.9894 \\
            \hline
            \multicolumn{2}{|c|}{Ours} & \textbf{31.14} & \textbf{0.9896} \\
            \hline
        \end{tabular}%
    }
\end{table}
As shown in Fig.~\ref{fig5}, we visualize the difference in image quality with and without internal- augmentation. The first and last columns represent the hazy images and ground truth images, respectively. The second column shows the dehazed images obtained using the external- augmentation method. The third column shows dehazed images obtained using both external data augmentor and the weak-to-strong augmentation self-supervised learning (WS-SSL) approach for internal data augmentor. As shown in Fig.~\ref{fig5}, by using the internal- augmentation and self-supervised learning (SSL), the details of the out-of-distribution image are closer to the real haze-free image. 

In Algorithm~\ref{alg:internal-augmentor}, we present the implementation of weak-to-strong self-supervised augmentation on internal data. We incorporate the self-supervised loss obtained from the internal dataset into the original loss. Therefore, we obtain the total loss from both external and internal data as follows:
\begin{equation}
\label{eq9}
\mathcal{L}_{total} =\mathcal{L}_{external}+\mathcal{L}_{internal}
\end{equation}

\section{Experiment}
\subsection{Experimental Settings}
Our model is implemented using PyTorch 1.13.0 on an NVIDIA RTX 3090 GPU. The model is trained using the AdamW optimizer~\cite{dehazeformer74} with a cosine annealing strategy~\cite{dehazeformer75}. The batch size and training epochs are set to 8 and 3000, respectively. The learning rate is reduced from an initial rate of 1e-4 to 1e-6. As shown in Fig.~\ref{fig12}, we present the loss curve of our method during the training process. It can be observed that the training curve converges when the epoch reaches 3000. Additionally, we apply the aforementioned external augmentation method to the RSID. Subsequently, we incorporate the revised RSID as an augmented dataset into the NID for training. 

\subsection{Dataset}
For dataset comparison, we select two datasets: The NID~\cite{trinityself},
and The RSID~\cite{trinityself}. For the NID, we select the first 900 images in the order of arrangement as the training set, while the remaining 100 images are used as the test set (N100). Therefore, for the RSCT 3:1 dataset, a total of 1200 images are used as the training set. For the RSCT 1:1 dataset, a total of 1800 images are used as the training set.

\subsection{Evaluation Metrics and Comparison Objects}
To evaluate the performance of our method, we use Peak Signal-to-Noise Ratio (PSNR), Structural Similarity Index (SSIM)~\cite{1284395}, Mean Squared Error (MSE), and Feature Similarity Index (FSIM)~\cite{5705575} as evaluation metrics. 

\subsection{Comparison with Baseline Methods}
We compare our method with state-of-the-art dehazing works. The experiments show that our results are significantly better than other methods, as shown in Table~\ref{table1} with specific experimental results.

Observing Table~\ref{table1}, we can see that our method first achieves a PSNR metric that reaches 31.14. In addition, our method significantly outperforms other approaches in all four evaluation metrics. We observe that compared to Trinity-Net~\cite{trinityself}, our method achieves a 2.34 dB improvement in PSNR, an increase of 0.0153 in SSIM and an improvement of 0.0056 in FSIM. Additionally, the MSE value after applying our method is reduced by 0.0608 compared to the Trinity-Net~\cite{trinityself}'s MSE of 0.2581. Compared to DEA-Net~\cite{deanet}, our method has improved by 3.1 dB in terms of PSNR and 0.0082 in terms of SSIM. Compared to UCL-Dehaze~\cite{ucldehaze}, our method demonstrates complete superiority across all four metrics. In contrast to MixDehazeNet~\cite{mixdehazenet}, our method achieves an improvement of 1.4 dB in the PSNR metric and an enhancement of 0.006 in SSIM. These results indicate that our method is better at restoring image visibility.

As shown in Fig.~\ref{fig6}, to specifically demonstrate the superiority of our method, we compared our model results with other methods, i.e., Zero-store~\cite{trinity18},UCL-Dehaze~\cite{ucldehaze} UHD~\cite{trinity13}, Dehamer~\cite{trinity21},  Trinity-Net~\cite{trinityself}, DEA-Net~\cite{deanet} and MixDehazeNet~\cite{mixdehazenet} . For example, from the images obtained above, it can be seen that our method approaches the real image more closely in terms of texture details. From the zoomed-in images, it can be observed that our dehazed images are noticeably closer to the real haze-free images in terms of texture and color. Moreover, the edge details are clearer compared to Trinity-Net~\cite{trinityself} and DEA-Net~\cite{deanet} methods.

We also apply our method to the remote sensing dataset (i.e., RSID). Fig.~\ref{fig7}, Fig.~\ref{fig8} specifically illustrates the comparison between our method and other methods. From the figures, it can be seen that the dehazed images obtained by Zero-Store~\cite{trinity18} and UCL-Dehaze~\cite{ucldehaze} still contain haze, while the dehazed images obtained by the UHD method~\cite{trinity13} exhibit partial distortion in terms of color. In comparison to DeHamer~\cite{trinity21}, in the results shown in the fourth column, there is still a small amount of residual haze along the edges. The results of  Trinity-Net~\cite{trinityself},DEA-Net~\cite{deanet} and MixDehazeNet~\cite{mixdehazenet} exhibit differences in local color compared to the haze-free ground truth image. In our method, the haze has been removed, and there is no distortion in color. The dehazed images obtained through our method are closest to the real haze-free images.

In Table~\ref{table2}, we observe that our method also significantly outperforms other approaches in all four evaluation metrics. For the Zero-Store~\cite{trinity18}, UHD~\cite{trinity13}, DeHamer~\cite{trinity21}, Trinity-Net~\cite{trinityself},UCL-Dehaze~\cite{ucldehaze}, MixDehazeNet~\cite{mixdehazenet} and DEA-Net~\cite{deanet} methods, we obtained actual quantitative results through reproduction. We utilize same experimental setup as reported in the Trinity-Net paper~\cite{trinityself}, ensuring consistency in the experimental conditions. Compared to Trinity-Net~\cite{trinityself}, our method achieved an improvement of 1.66 dB in the quantitative metric PSNR and an improvement of 0.0175 in SSIM. Compared to UCL-Dehaze~\cite{ucldehaze}, our method also shows comprehensive improvements across all four metrics. In comparison to MixDehazeNet~\cite{mixdehazenet}, our method achieves an improvement of 1.24 dB in the PSNR metric and an enhancement of approximately 0.016 in SSIM. These quantitative results indicate that our method also produces dehazed images with higher visual quality on the remote sensing dataset RSID.

\emph{Cross-model evaluation.} To further evaluate our algorithm, we apply our method to both Trinity-Net and DEA-Net to better demonstrate the generalizability across other models. As shown in Table~\ref{table7} and~\ref{table8}, on the NID, by applying our method, the PSNR of the NID test set increases by 0.32 dB, and the SSIM improves by 0.0012. Compared to the original data augmentation method, using our method increases the PSNR of the RSID test set by approximately 0.5 dB and the SSIM by 0.0039. This demonstrates that our method is also applicable to other models and enhances the robustness of the models.

\emph{Comparison with the full supervised framework.} To further verify the effectiveness of our self-supervised learning method, we compare it with the method using fixed ground truth $y$. As shown in Table~\ref{table9}, our method achieves a PSNR improvement of 0.28 dB and an SSIM improvement of 0.0002 on the NID test set compared to the supervised framework (using fixed $y$). The experimental results demonstrate that the dynamically generated $I_{week}$ and $I_{aggr}$ outperform the fixed label supervision in capturing local structural features. This mechanism encourages the network to learn robust representations from its own outputs, indicating that our method more effectively integrates the dynamic predictions of the model during training.

\textbf{Efficiency analysis.} As shown in Table~\ref{table6}, we conduct an efficiency analysis of our method compared to other approaches. Our method is evaluated based on the NID (1:1) dataset. Although our method has higher runtime and computational complexity (GFLOPS), the number of parameters in our model is significantly lower than that of other methods. Compared to Trinity-Net~\cite{trinityself} and DEA-Net~\cite{deanet}, our method maintains the highest evaluation metrics while having fewer parameters than both. The experimental results confirm that our method offers high accuracy and practical usability.

\subsection{Ablation Study}
To demonstrate the effectiveness of our proposed method, we conduct ablation studies to analyze the effects of each module.

\emph{Effectiveness of internal-augmentation.} In ablation study, we use L1 loss as the initial image reconstruction loss (Eq.~\ref{eq4}), while in the self-supervised method, we use $\mathcal{L}_{internal}$ loss. The remaining one hundred images from the NID dataset (NID100) are used as the test set. In Table~\ref{table3}, we only use internal- augmentation and a self-supervised module in our method. Compared to not adding modules, using internal- augmentation and self-supervised modules results in an improvement of 0.14 dB in PSNR values on NID100. In Fig.~\ref{fig9}, partial dehazed images obtained after internal- augmentation and self-supervised modules are shown.

\emph{Effectiveness of external-augmentation.} In Table~\ref{table4}, we solely employ external- augmentation in the model. Specifically, we incorporate external- augmentation by including the augmented RSID dataset in the training set. The training images from the NID and the RSID are divided into a ratio of 3:1 and 1:1, respectively, for training. In Table~\ref{table4}, compared to the results obtained without adding any modules, the results obtained with a 1:1 ratio show improvements of 0.65dB and 0.0021 in PSNR and SSIM metrics, respectively. Compared to the 3:1 ratio, the 1:1 ratio shows improvements of 0.24 and 0.0014 in PSNR and SSIM metrics, respectively. RSCT 3:1 only expanded 300 gamma-corrected images, resulting in a total of 1200 images, while RSCT 1:1 expanded 900 gamma-corrected images, resulting in a total of 1800 images. As a result, RSCT 3:1 has a smaller dataset size and less data diversity compared to RSCT 1:1. This leads to the model learning fewer features during training than RSCT 1:1, reducing the model's robustness and consequently resulting in lower performance on the test set compared to the results obtained from training with RSCT 1:1. Additionally, Fig.~\ref{fig10} shows the experimental results of haze removal with external- augmentation at ratios of 3:1 and 1:1, respectively.

\emph{Effectiveness of external- and internal- augmentation.} Then, we train the network using both external and internal augmentation methods simultaneously, as shown in Table~\ref{table5}. We observe that the highest PSNR value of 31.14dB is achieved when simultaneously using the weak-strong augmentation self-supervised method and dividing the training data of the NID and the RSID in a 1:1 ratio during the channel transfer module. Compared to the results obtained without adding any modules, the PSNR value on the test set has increased by 1.08dB, and the SSIM has increased by 0.0025. Compared to the 3:1 split of training data between the NID and the RSID, the PSNR on the test set obtained through the 1:1 split has increased by 0.44dB. As shown in Fig.~\ref{fig11}, the single dehazed image obtained after applying both modules exhibits higher visual quality and is closer to the true haze-free image in terms of overall color.

\section{Conclusion and Limitation}
In this paper, we propose a novel method that performs data augmentation separately on the external and internal aspects of the data. Specifically, we achieve external- augmentation and internal- augmentation using weak-to-strong self-supervised learning. Our method leverages data diversity at the external level and further explores internal information to effectively dehaze images. 

Although our method achieves superior results in both quantitative and qualitative evaluations, the introduction of augmentation operations has led to an increase in training time compared to other methods. In the future, we will focus on research aimed at improving training efficiency.

\bibliographystyle{IEEEtran}
\bibliography{ref}

\end{document}